\documentclass[conference]{IEEEtran}
\IEEEoverridecommandlockouts

\usepackage{booktabs}
\usepackage{tikz}
\usepackage{circuitikz}
\usepackage{soul}

\newcommand\copyrighttext{%
  \footnotesize \textcopyright 2024 IEEE. Personal use of this material is permitted.  Permission from IEEE must be obtained for all other uses, in any current or future media, including reprinting/republishing this material for advertising or promotional purposes, creating new collective works, for resale or redistribution to servers or lists, or reuse of any copyrighted component of this work in other works.}
  
\newcommand\copyrightnotice{%
\begin{tikzpicture}[remember picture,overlay]
\node[anchor=south,yshift=10pt] at (current page.south) {\fbox{\parbox{\dimexpr\textwidth-\fboxsep-\fboxrule\relax}{\copyrighttext}}};
\end{tikzpicture}%
}

\begin{document}

\title{SAGA: Synthesis Augmentation with Genetic Algorithms for In-Memory Sequence Optimization\\
\thanks{Identify applicable funding agency here. If none, delete this.}
}

\author{\IEEEauthorblockN{Andey Robins and Mike Borowczak}
\IEEEauthorblockA{\textit{Department of Electrical and Computer Engineering} \\
\textit{University of Central Florida}\\
Orlando, USA \\
\{andey.robins, mike.borowczak\}@ucf.edu}}
% blinded@blin.ded}}

\maketitle
\copyrightnotice
\begin{abstract}
  The von-Neumann architecture has a bottleneck which limits the speed at which data can be made available for computation. To combat this problem, novel paradigms for computing are being developed. One such paradigm, known as in-memory computing, interleaves computation with the storage of data within the same circuits. MAGIC, or Memristor Aided Logic, is an approach which uses memory circuits which physically perform computation through write operations to memory. Sequencing these operations is a computationally difficult problem which is directly correlated with the cost of solutions using MAGIC based in-memory computation. SAGA models the execution sequences as a topological sorting problem which makes the optimization well-suited for genetic algorithms. We then detail the formation and implementation of these genetic algorithms and evaluate them over a number of open circuit implementations. The memory-footprint needed for evaluating each of these circuits is decreased by up to 52\% from existing, greedy-algorithm-based optimization solutions. Over the 10 benchmark circuits evaluated, these modifications lead to an overall improvement in the efficiency of in-memory circuit evaluation of 128\% in the best case and 27.5\% on average.
\end{abstract}


\maketitle

\section{Introduction}

Data and the locations where it is used in computation are physically separated in modern computing architectures. Data is repeatedly moved physically nearer to where computations are performed before being used for whatever purpose it was retrieved. This paradigm of computer architecture has been instrumental in the development of modern computers and the ways we utilize them. As processing has sped up though, this transfer of information from storage to compute has formed a bottleneck which limits the computational speeds which can be achieved by state-of-the-art devices. Specialized processing units, such as Graphics Processing Units (GPUs) and other purpose-built hardware, often attempt to side-step the problem by increasing the throughput of information; however, the theoretical problem of moving data around is not solved by this approach; it is only mitigated in particular situations. Additionally, specialized processors are often tailored to a specific domain. As an example, GPUs excel at matrix and vector operations, but their strengths are under-utilized for highly sequential tasks. 

An alternative paradigm to the von-Neumann architecture would be performing the computation at the same place the data is stored rather than transporting the data to another location for processing. This computing paradigm is aptly referred to as ``processing in-memory" (PIM). Memristor Aided Logic (MAGIC) is an emerging PIM paradigm making use of parallel, write-based operations to perform calculations in-memory \cite{kvatinsky2014magic}. Updating the foundational architecture of a computing system has many abstract and novel challenges. One such problem is that performing a calculation requires scheduling operations for the computation; however, the number of in-memory cells needed to evaluate the function described is dependent on the scheduling order. State-of-the-art solutions model this dependence as a graph and seek to solve the scheduling challenge as a covering problem. In this work, we instead re-frame this problem as one of permutation, applying that idea to the development of a genetic algorithm (GA) which produces reductions in the memory footprint of execution up to 52.8\% in the best case when compared to the prior state-of-the-art on standard benchmark circuits \cite{ben2019simpler}.

This work proposes an additional synthesis phase for area optimization of PIM circuits, referred to as SAGA or Synthesis Augmentation with Genetic Algorithms, which integrates a machine-learning framework for final optimizations and is organized as follows. Section \ref{sec:prior-works} contextualizes this work by describing prior systems for optimizing the cost of MAGIC-based in-memory circuits before using them as a benchmark for comparison of memory footprints. Section \ref{sec:methods} describes both the design of the evolutionary framework utilized to improve operation sequencing and also describes the experiments used to evolve more fit solutions. Section \ref{sec:results} presents metrics and Section \ref{sec:discussion} discusses the improvement on benchmark circuits.

\section{Prior Work}\label{sec:prior-works}

SAGA seeks to combine GAs with a CAD workflow, and begins by contextualizing PIM computation as one of many emerging paradigms for computing. We discuss MAGIC, a specific PIM logic style, and recent frameworks for optimizing circuit synthesis for the technology underpinning it, before providing a brief description of the fundamental pieces of the GAs used. 

\subsection{Emerging Paradigms}

Limitations of contemporary computing architectures have led to the emergence of novel computing paradigms which work to overcome the limitations of Dennard scaling and the slowing of Moore's Law \cite{johnsson2016impact, haensch2017scaling}. Quantum computing \cite{gyongyosi2019survey} may be the most well known alternative paradigm for computing, but others, though less well-known, approach the problem of novel architecture in various ways. Photonic computing \cite{xiang2021review}, analog computing \cite{zangeneh2021analogue}, and PIM \cite{verma2019memory} are examples of emerging solutions which all diverge from current architectures in meaningful ways. 

As an emerging paradigm, in-memory computation can itself be subdivided into different categories. An entire field of study currently quantifies applications of analog circuits for performing computation \cite{soliman2020ultra}; and special focus has been recently paid towards applications of this domain for artificial intelligence applications \cite{antolini2023combined, rasch2023hardware}. Digital in-memory computation though is preferable for many applications due to the approximations implicit to analog computation. These digital paradigms are often separated by whether they are parallel or non-parallel as well as if they are read or write based \cite{rashed2022stream, zha2016reconfigurable}. SAGA specifically targets non-parallel operations using write-based PIM for area optimization.

\subsection{Processing In-Memory}

MAGIC performs computation by loading a memristor with a logical value and applying voltage across the circuit. The memory value stored by this operation is then dependent on the value which was loaded, allowing for the construction of simple gates. By combining memristors, more complex circuits, such as NOR gates of various input sizes, are formed. Individual memristors are then wired into a grid structure known as a crossbar matrix \cite{rashed2021hybrid} allowing for some CAD tools to make use of writes to evaluate multiple NOT operations in parallel.

\begin{figure}[!ht]\label{fig:footprint}
    \centering
    \hspace{-2em}
    \includegraphics[width=0.4\textwidth]{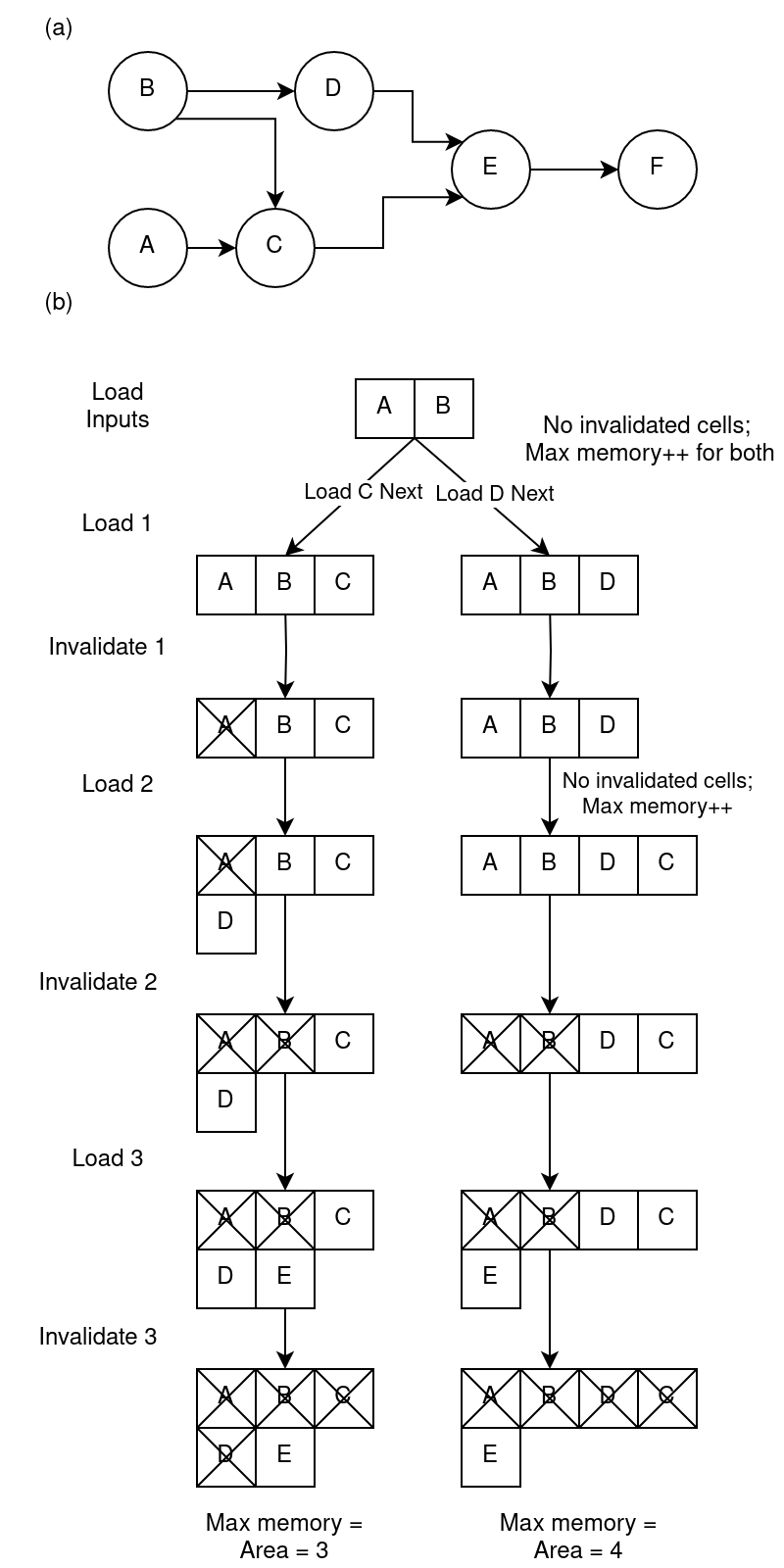}
    \caption{
    Sub-figure (a) details an execution graph with memory cell costs depending on the evaluation order of the graph. Two possible orders exist which fit the constraints described in Section \ref{sec:problem-formulation}. Processing vertex C before D has cost 3 while processing vertex D before C has cost 4 as illustrated in sub-figure (b).}
    \label{fig:enter-label}
\end{figure}

With respect to MAGIC, application of formal design principles has continued with advances in technology mapping directly improving either the number of cycles used to complete an operation, the number of memory cells required for the operation, or both, which improves the efficiency of the PIM circuit overall \cite{rashedautomated}. Since evaluating a function using MAGIC requires the translation of a netlist into a sequence of in-memory memristor operations, this sequencing has significant effects on the amount of memory required to evaluate a function. Figure 1 illustrates this trade-off in the execution order. When the graph is processed in the ordering $A\rightarrow B\rightarrow C\rightarrow D\rightarrow E\rightarrow F$, only three memory cells are required. When the graph is processed $A\rightarrow B\rightarrow D\rightarrow C\rightarrow E\rightarrow F$, it requires four memory cells. For a graph with only six vertices, swapping the processing order of two nodes in the graph directly improves the efficiency without changing any behavior of the circuit. 

A number of frameworks and processes for simplifying the execution sequences for MAGIC have been introduced. One of these early frameworks was the \textsc{Simple} synthesis tool \cite{hur2017simple}. This tool provided a framework for CAD workflows to be translated into the realm of in-memory computation. This  process first used the tool \texttt{abc} to minimize circuits in a technology agnostic manner \cite{abc}. Then, the framework performed an additional round of technology dependent optimization to exploit the specific constraints of MAGIC systems resulting in nearly halving the best execution costs at the time.

Another approach attempting to develop a framework for this process applied a look-ahead operation to optimize the sequencing of operations \cite{yadav2019look}. Based on the fan-out for each gate, the algorithm proposed was able to improve the overall number of cycles required for an operation. Their work follows a similar synthesis flow to \textsc{Simple} by applying their optimization approach to the synthesis output of \texttt{abc} before final technology mapping. This work-flow, of offloading initial circuit simplification to \texttt{abc}, is a process mirrored by the process detailed in Section \ref{sec:methods}.

Eventually, \textsc{Simple} was followed by \textsc{Simpler} which made use of a metric referred to as ``cell usage" \cite{ben2019simpler} which can be calculated from the output of \texttt{abc}. The cell usage of each gate enabled a search strategy which once again reduced the execution cost of in-memory circuits substantially. Across multiple metrics, there were improvements such as a 63x average increase in area efficiency, a 10\% decrease in the number of operations, and a 5x decrease in the latency of parallel operations.

\begin{figure*}[!ht]
    \centering
    \includegraphics[height=0.25\textwidth]{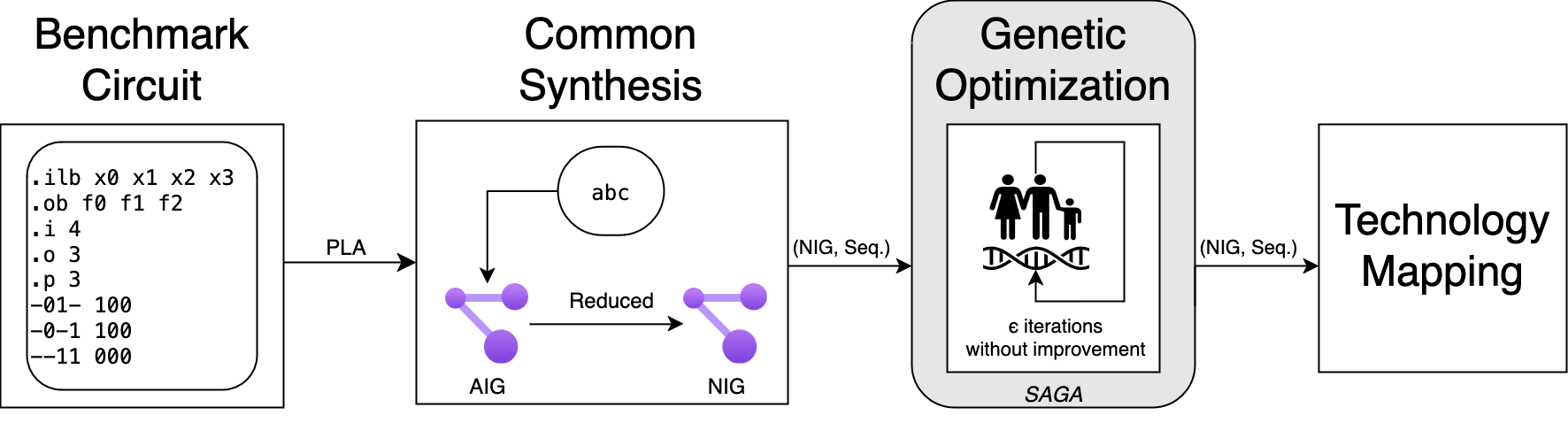}
    \caption{The synthesis flow from benchmark circuit specifications to genetically optimized execution sequences.}
    \label{fig:gene-flow}
\end{figure*}

\subsection{Genetic Algorithms}\label{sec:gas}

Advances in the field of CAD and artificial intelligence (AI) have been tied historically \cite{jiaoying1987artificial}, but the applications of AI and CAD to emerging computing paradigms has been more limited. Often the design and fabrication of other physical components through processes such as 3D printing \cite{hunde2022future} are the focus. Therefore, a secondary offering of this work is a demonstration of the feasibility of applying AI techniques within the CAD pipeline for circuit synthesis. GAs in particular have been applied to similar sequencing and scheduling problems in the CAD space \cite{squillero2011artificial}, but not for PIM CAD. 

GAs are an effective tool for problems where a large population of candidate solutions are able to be evaluated, and small changes between individuals can be quickly made. Sequence optimization problems are one such framing for which GAs are known to be effective. Historically, there has been more exploration of genetic evolution algorithms within the space of "pseudo-Boolean optimization problems," but sequence optimization has also benefited from these GAs \cite{doerr2022towards}.

A GA is composed of four elements: a means by which a population of individuals can be evaluated and sorted by efficacy (evaluation), a method to select a subset of the population to seed the next population (selection), a means to combine individuals to create more (reproduction), and methods to re-integrate and update individuals in the population (replacement) \cite{sivanandam2008genetic}. These individual components lead to an emergent behavior of optimizing the fitness function used in evaluation to find the most fit individual. For evolutionary problems which can be modeled as graphs, such as sequencing operations, the relationships between descendants of nodes and the graph at large are significantly affected by mutation operators \cite{allen2012mutation}.

\section{Methods}\label{sec:methods}

The end-to-end process in SAGA duplicates the initial processes used by prior MAGIC synthesis tools like \textsc{Simple} and \textsc{Simpler}. This is detailed in Figure \ref{fig:gene-flow}. SAGA introduces an additional step following this common synthesis pass which applies GAs to the domain of PIM circuit synthesis optimization for the first time to our knowledge. Since this optimization occurs at the end of netlist synthesis and before technology mapping, SAGA may be integrated into other in-memory synthesis workflows which begin with common logical formula specifications (e.g. BLIF, PLA, etc.).

\subsection{Problem Formulation}\label{sec:problem-formulation}

For a given graph, each topological sort of the vertices corresponds directly with a sequence of PIM operations. Depending on the sequence of these operations, a variable number of memory cells are required to include all of the intermediary products in the computation; and finding a sequence to optimize the number of cells is an NP-Hard problem \cite{sethi1973complete}. For a given sequence, the fitness of that sequence is the number of memory cells its execution would require. The problem is then realized as a minimization of the fitness function. To validate the efficacy of SAGA, we test the null hypothesis $H_0$: There is no statistically significant difference in the circuit synthesis between \textsc{Simpler} and SAGA for the benchmark circuits.

Previous works have applied traditional CAD synthesis flows to optimizing this problem. They have applied tools like \texttt{abc} \cite{abc} to minimize the logical expressions of a function and applied greedy algorithms to cover the resulting graph with circuit components. Novel approaches to framing the problem, such as decomposing the calculations into larger intermediary components \cite{rashedautomated}, provide ways to reduce higher-level problems to the technological tools currently available. By reformulating the synthesis problem, this work instead structures the problem as one which can be tackled with genetic evolution and then applies evolutionary algorithms as a final optimization step in the synthesis work flow.

A circuit specification is taken from RevLib \cite{revlib} and loaded into \texttt{abc} \cite{abc}. The circuit is then internally reduced by \texttt{abc} and mapped to a NOR-Inverter Graph, or NIG, using a custom circuit library as in \textsc{Simpler} \cite{ben2019simpler}. This graph is then parsed into the genetic evolution algorithm and the population is optimized using a generational GA. Finally, an optimized execution sequence is output by SAGA to be incorporated into the final technology mapping and execution of the in-memory function. We describe the process by which the netlist produced by \texttt{abc} is translated to the GA domain and the genetic operators utilized in SAGA. Finally, we describe the configuration of the experimental population and stopping conditions for synthesis. To evaluate $H_0$, we employ a one-tailed t-test to determine if the differences in circuit efficiency are statistically significant with significance levels set at $\alpha=0.05$.

\subsection{Netlist Transcription}\label{sec:mapping}

Benchmark circuits from RevLib are initially specified in Programmable Logic Array (PLA) form \cite{revlib}. The PLA encoding is then converted, synthesized and mapped to a minimal technology library of inverter and 2-input NOR gates by \texttt{abc}. The resulting NIG is then converted into a DAG with edges oriented in the direction of data flow. The graph is topologically sorted into a list of vertices corresponding to an execution sequence for the gates in the circuit using a breadth first traversal of the DAG beginning with all input vertices being visited first. Other approaches in the literature perform their optimizations at this sorting phase; this work, uses a computationally cheap sorting which produces a sub-optimal solution which is used to initialize the population for genetic optimization.

\subsection{Genetic Operators}\label{sec:genetic-ops}

Crossover is performed with a single point, order based crossover \cite{deep2011new}. One chromosome is read up until the mutation point, then beginning at the mutation point, all sequence values are read and added if they are not already on the new chromosome. Mutation is performed by selecting a point on the sequence and swapping it with another value which preserves the validity of the sequence. Determining valid swap candidates can be done by filtering out all nodes which are directly dependent upon or descended from the currently mutating vertex. This can be efficiently computed once as the relationship of vertices in the graph doesn't change during genetic evolution.

\subsection{Genetic Experiment}\label{sec:genetic-experiment}

As SAGA presents an application of GAs as a final step in optimization, the online performance of the algorithm is of less importance than the final, offline fitness. As such, the hyperparameters are tuned with the mindset that larger populations which take longer to evaluate and combine are acceptable provided the algorithm produces execution sequences with minimal memory footprint. Additional work would be necessary to determine for which hyperparameters the synthesis time is minimized while preserving performance.

Populations of 2000 individuals are initially created with a 20\% chance of a sequence being mutated once during each recombination. Sequences have their fitness evaluated by simulating each sequence in memory and recording the footprint. Simulation makes use of a mark-and-sweep algorithm to ensure that the minimal number of memory cells are utilized for each evaluation. 

Following fitness evaluation, the half of the population which are least fit are discarded. Individuals are then sorted according to their fitness, paired with another individual next to them in sorted order, and undergo the ordered crossover operation previously described. After crossover, each individual in the population will have the mutation operator applied with probability 20\%. Generations are evaluated until $\epsilon$ generations pass with no improvement to the fitness of the best individual in the population. Multiple values of $\epsilon$ are evaluated and discussed in Table III. The best observed sequence across all runs is reported in Table I.

\begin{table*}[!ht]
    \centering
    \label{tab:performance}
    \caption{Performance comparison between this work and the best performing prior works. Efficiency is calculated according to the formula Efficiency $ = 10^6 \cdot Area^{-1} \cdot Cycles^{-1}$ \cite{ben2019simpler}. Efficiency metrics have been rounded to the nearest whole number for display where greater is better.} 
    \begin{tabular}{@{}l|ccc|ccc|cccc|ccc@{}}
    \toprule
         & \multicolumn{3}{c|}{\textsc{Simple MAGIC} \cite{hur2017simple}}& \multicolumn{3}{c|}{\textsc{Simpler MAGIC} \cite{ben2019simpler}} &\multicolumn{4}{c|}{SAGA [This Work]}& \multicolumn{3}{c}{Improvement} \\ \midrule
        Benchmark  & Cycles& Area&Efficiency& Cycles & Area & Efficiency & $\epsilon$&Cycles  & Area & Efficiency & Cycles & Area & Efficiency \\ \midrule
         % & & & & & & & & & \\
        \texttt{5xp1}  & 886& 315&4& 119 & 39 & 215 & 5000&160  & 29& 216& -34.5\% & 25.6\%& 0.47\%\\
        \texttt{9symml}  & n/a& n/a&n/a& 218 & 57 & 80 & 5000&306  & 57& 57& -40.4\%& 0.0\%& -28.8\%\\
        \texttt{clip}  & 742& 444&3& 160 & 47 & 133 & 5000&233 & 40& 107& -45.6\% & 14.9\%& -19.5\%\\
        \texttt{cm150a}  & 570& 189&3& 67 & 39 & 383 & 50&52  & 22 & 874& 22.4\% & 43.6\%& 128\%\\
        \texttt{cm162a}  & 530& 186&9& 64 & 35 & 446 & 50&87 & 20& 575& -35.9\%& 42.9\%& 28.9\%\\
        \texttt{cm163a}  & 522& 183&10& 66 & 36 & 421 & 50&76  & 17 & 774& -15.2\% & 52.8\%& 83.8\%\\
        \texttt{misex1}  & 1380& 294&2& 83 & 33 & 365 & 500&84  & 17 & 700& -1.20\% & 48.5\%& 91.8\%\\
        \texttt{parity}  & 1078& 240&4& 81 & 35 & 353 & 500&104 & 20& 481& -28.4\%& 42.9\%& 36.3\%\\
        \texttt{sao2}  & n/a& n/a&n/a& 128 & 53 & 147 & 5000&213  & 43 & 109& -66.4\%& 18.9\%& -25.9\%\\
        \texttt{x2}  & 1404& 168&4& 73 & 33 & 415 & 5000&80 & 16& 781& -9.59\%& 51.5\%& 88.2\%\\
        \bottomrule
    \end{tabular}
\end{table*}

\section{Results}\label{sec:results}

Unless otherwise specified, averages refer to the geometric mean of a value for all benchmarks. Compared to the previous state-of-the-art, using SAGA to optimize MAGIC-based circuits leads to area improvements of up to 52.8\% and cycle improvements of up to 22.4\% with overall improvements in the total efficiency of the benchmark circuits up to 128\%. Based on the data collected, the null-hypothesis, $H_0$ with an $\alpha=0.05$ and a resulting p-value of 0.02934, is rejected. Thus the improvements from SAGA are significant.

\begin{table}
    \centering
    \label{tab:statistics}
    \caption{The average percentage change of each statistic where negative values are decreases in the metric while positive are improvements.} 
    \begin{tabular}{l|ccc}
    \toprule
        Statistic & Cycles & Area & Efficiency \\
        \midrule
        Arithmetic Mean& -25.5\%& 33.6\%& 38.3\%\\
        Geometric Mean& -29.5\%& 32.3\%& 27.5\%\\
        Standard Deviation& 25.3\%& 19.2\% & 56.7\%\\
        95\% Confidence&(-41.1, -9.82) & (21.7, 45.5)& (3.17, 73.5) \\
    \bottomrule
    \end{tabular}
\end{table}

On average, the application of genetic evolution to the execution sequence was responsible for a 32.3\% decrease in the number of memory cells required for each circuit compared to \textsc{Simpler}. Only one circuit didn't have the number of memory cells required reduced with this method, namely the \texttt{9symml} benchmark. In regards to the number of cycles required for each circuit, unlike with \textsc{Simpler} which reduced the cycle count by executing NOT operations in parallel, no operations are performed in parallel in this work. As a result, the number of cycles executed for each circuit is 29.5\% higher on average than it was with \textsc{Simpler}. Despite this, one benchmark had a reduction in the number of cycles required by synthesizing a smaller circuit at the \texttt{abc} phase of synthesis. In the worst case, cycle counts were increased by 66.4\% compared to \textsc{Simpler}. The implications of using purely sequential operations within this work are explored in further detail in Section \ref{sec:discussion}. The overall efficiency improvements were 27.5\% on average for the benchmark circuits. More complete statistics including 95\% confidence intervals on the impact of the changes is presented in Table II. 

A common limitation of GAs is the computational effort required to generate highly optimized solutions can significantly exceed the effort required to find moderately optimized solutions. In an effort to demonstrate and quantify the impact of longer search time, the performance for various values of epsilon was also examined and presented in Table III.

\section{Discussion}\label{sec:discussion}

In contrast to one of the prior state-of-the-art works no additional work to identify operations which can be performed in parallel is done. Even without these additional optimizations the efficiency increased in the majority of cases and on average for the benchmark circuits. The memory footprint, the primary target of optimization in this work, was constant or improved in all benchmark cases.

Reasons for not including a parallel optimization process are three-fold. First, as a demonstration that this technique is capable of improving the efficiency of PIM circuits with less specialized knowledge than hand-crafted CAD algorithms provides support to the argument that applying GAs for optimization of PIM operations is an effective choice. Secondly, modifying the fitness function to maximize efficiency rather than minimizing area could potentially improve the exact measures of efficiency, but would not fundamentally change the approach to the problem, which is the primary contribution of this work. Additionally, while NOT operations in the MAGIC paradigm can be simply parallelized due to the structure of MAGIC crossbars utilized in the literature, this structure may not necessarily fit for all PIM operation sequencing problems; however, the genetic evolution approaches to evaluating this synthesis and integrating it into the design work-flow could still be applied in a way to minimize the cost of physical implementation for subsequent technologies.

Due to the stage of the process in which this genetic evolution-backed optimization occurs, validation of the evolved sequence is a computationally cheap operation. In fact, it's performed during fitness calculations to ensure solutions are valid throughout the search. This is accomplished by verifying the sequence can visit every node in the NIG without visiting a node whose parents have not been visited. 

\begin{table*}[!ht]
\label{tab:epsilon}
    \centering
    \caption{Performance comparison for various epsilon. The percentage change from one value of $\epsilon$ to the next is presented in the $\Delta_{eff}$ column.} 
    \begin{tabular}{@{}l|c|cc|ccc|ccc@{}l}
    \toprule
         \multicolumn{2}{c|}{}&\multicolumn{2}{c|}{$\epsilon = 50$}&\multicolumn{3}{c|}{$\epsilon = 500$}&\multicolumn{3}{c}{$\epsilon = 5000$}\\ \midrule
        Benchmark  & Cycles & Area & Efficiency & Area & Efficiency  &$\Delta_eff$& Area & Efficiency  &$\Delta_eff$\\ \midrule
        \texttt{5xp1}  & 160& 37& 169& 30& 208 &23.1\% & 29& 216 & 3.85\%\\
        \texttt{9symml}  & 306& 66& 50& 63& 52 & 4.00\% & 57& 54 & 9.52\%\\
        \texttt{clip}  & 233& 50& 86& 43& 100 & 16.3\% & 41& 105 & 5\%\\
        \texttt{cm150a}  & 52& 22& 874& 22& 874 & 0\% & 22& 874 & 0\%\\
        \texttt{cm162a}  & 87& 21& 547& 21& 547 & 0\%& 20& 575 & 0\% \\
        \texttt{cm163a}  & 76& 17& 774& 17& 774 & 0\%& 17& 774 & 4.76\%\\
        \texttt{misex1}  & 84& 18& 661& 17& 700 & 5.90\%& 17& 700 & 0\%\\
        \texttt{parity}  & 104& 25& 385& 20& 481 & 24.9\%& 20& 481 & 0\%\\
        \texttt{sao2}  & 213& 47& 100& 44& 107 & 7\%& 43& 109 & 1.87\%\\
        \texttt{x2}  & 80& 19& 658& 17& 735 & 11.7\%& 16& 781 & 6.26\%\\ \bottomrule
    \end{tabular}
\end{table*}  

\section{Conclusion}\label{sec:conclusion}

GAs are effective optimization tools at the final stage in the design process. Future work would be needed to assess the integration of these algorithms into earlier phases of the design workflow or the replacement of traditional design algorithms with machine learning algorithms such as this one. Furthermore, whether additional tuning to evolution hyperparameters and the fitness function lead to further improvements is an open question which would require more statistical analyses than this work is able to present. Finally, while the time needed to synthesize an individual circuit is not a primary focus of this work, in a production setting, expeditious synthesis is a valuable property. Modifications to the synthesis flow which lead to faster convergence may be desirable for applications of this technology. Further work would be necessary to explore the impacts to synthesis time caused by modifications to the configuration of the genetic evolution workflow.

This work presents a workflow for applying genetic evolution algorithms to the synthesis of in-memory memristor-based operation sequences for realizing boolean functions. The injection of machine learning algorithms into the CAD pipeline allows for final optimizations to be conducted beyond the current state of the art, human-crafted algorithms and is made available on GitHub \cite{saga}. By reformulating the synthesis of circuits problem from the more common framing as a covering problem to a problem of sequence permutation, GAs shine as the machine learning tool capable of optimizing the final footprints of circuits. Using SAGA leads to reductions in the number of in-memory memory cells needed for in-memory computing by 32\% and overall improvements in efficiency of 27\% on average across the evaluated benchmark circuits.

\newpage

\section*{Acknowledgment}
This work was funded in part by the ORCGS fellowship at the University of Central Florida. Furthermore, this work extends prior work from a project for the course EEE 5336 on CAD of VLSI at the University of Central Florida in Fall of 2023.

\bibliographystyle{IEEEtran}
\bibliography{sample-base}

\end{document}